\documentclass[letterpaper,10pt,conference]{ieeeconf}
\IEEEoverridecommandlockouts
\usepackage[T1]{fontenc}
\usepackage[utf8]{inputenc}
\usepackage{times}
\usepackage{amsmath,amssymb}
\usepackage{graphicx}
\usepackage{booktabs}
\usepackage{flushend}
\usepackage[nocompress]{cite}
\usepackage{xurl}
\usepackage[hidelinks]{hyperref}
\AtBeginDocument{%
  \setlength{\abovedisplayskip}{4pt plus 1pt minus 1pt}%
  \setlength{\belowdisplayskip}{4pt plus 1pt minus 1pt}%
  \setlength{\abovedisplayshortskip}{0pt plus 1pt}%
  \setlength{\belowdisplayshortskip}{4pt plus 1pt minus 1pt}%
  \setlength{\textfloatsep}{12pt plus 2pt minus 2pt}%
  \setlength{\dbltextfloatsep}{14pt plus 2pt minus 2pt}%
  \setlength{\intextsep}{10pt plus 2pt minus 2pt}%
}
\makeatletter
\renewcommand{\section}{\@startsection{section}{1}{\z@}
  {1.5ex minus 0.5ex}{0.7ex}{\normalfont\normalsize\centering\scshape}}
\renewcommand{\subsection}{\@startsection{subsection}{2}{\z@}
  {1.5ex minus 0.5ex}{0.7ex}{\normalfont\normalsize\itshape}}
\makeatother
\AddToHook{env/thebibliography/begin}{\interlinepenalty=10000}
\newcommand{\method}{Perceptive SONIC}
\newcommand{\R}{\mathbb{R}}
\newcommand{\sg}{\operatorname{sg}}
\title{\LARGE\bf Perceptive SONIC: Terrain-Aware Residual Adaptation\\of a Frozen Behavior Foundation Model}
\author{Zikun Zhuang$^{1,2,\dagger}$, Ming Lei$^{2,3,\dagger}$, Jun Wang$^{2}$, Malin Wang$^{2,*}$,\\
and Zhiyun Lin$^{1,*}$, \IEEEmembership{Fellow, IEEE}%
\thanks{$^{1}$Zikun Zhuang and Zhiyun Lin are with Southern University of Science and Technology, Shenzhen, China.\newline E-mail: {\tt\footnotesize linzy@sustech.edu.cn} (Zhiyun Lin).}%
\thanks{$^{2}$Zikun Zhuang, Ming Lei, Jun Wang, and Malin Wang are with OPPO Research Institute, Shenzhen, China.\newline E-mail: {\tt\footnotesize wangmalin123@gmail.com} (Malin Wang).}%
\thanks{$^{3}$Ming Lei is with Shandong University, Jinan, China.}%
\thanks{$^{\dagger}$This work was done while Zikun Zhuang and Ming Lei were interns at OPPO Research Institute.}%
\thanks{$^{*}$Corresponding authors: Malin Wang and Zhiyun Lin.}%
\thanks{This work has been submitted to the IEEE for possible publication. Copyright may be transferred without notice, after which this version may no longer be accessible.}%
}
\hypersetup{pdfauthor={Zikun Zhuang, Ming Lei, Jun Wang, Malin Wang, and Zhiyun Lin},pdftitle={Perceptive SONIC: Terrain-Aware Residual Adaptation of a Frozen Behavior Foundation Model}}
\begin{document}
\bstctlcite{bibcontrol}
\IEEEaftertitletext{% Full-width first-page teaser, inserted through the class title hook.
\begin{minipage}{\textwidth}
    \centering
    \includegraphics[width=0.95\textwidth]{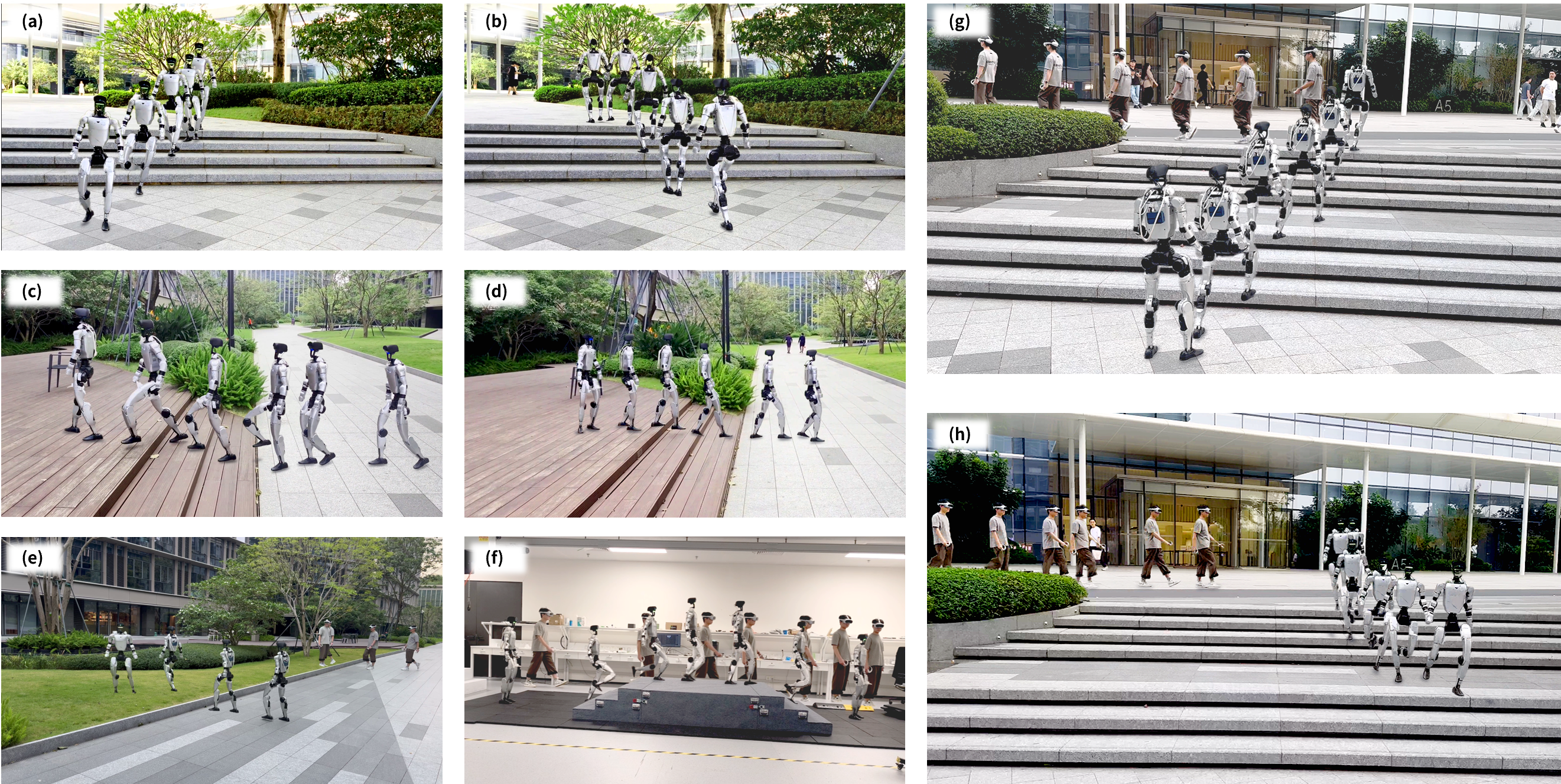}
    \makeatletter\def\@captype{figure}\makeatother
    \caption{Perceptive SONIC adapts flat-ground motions to complex terrain. SONIC Planner: (a) stone-step descent, (b) stone-step ascent, (c) wooden-step ascent, and (d) wooden-step descent. VR teleoperation: (e) grass walking; (f) indoor stair traversal; (g) stone-step descent; (h) stone-step ascent.}
    \label{fig:hardware_deployment}
    \vspace{8pt}
\end{minipage}
}
\maketitle
\thispagestyle{empty}
\pagestyle{empty}
% Align column bottoms without stretching section-heading gaps.
\flushbottom
\begin{abstract}
Humanoid behavior foundation models (BFMs) provide coordinated whole-body behavior, but terrain-independent references can prescribe contacts that are incompatible with the current scene. We present Perceptive SONIC, a perceptive joint-action residual framework that reuses off-the-shelf SONIC. This avoids both constructing paired raw/terrain-adapted motion datasets and training an additional tracking prior. A privileged residual teacher retrieves spatial terrain features through separate foot queries and gated cross-attention. Reference-contact annotations supervise stance support and penalize mistimed, insufficient, or repeated contacts, while a swept-clearance reward encourages foot clearance over terrain along upcoming reference-foot paths. Online distillation transfers residual-action means and spatial tokens to a depth-based student. In simulation evaluations spanning diverse motions and terrains, our framework improves success from 58.1\% for SONIC to 91.6\%, and from 43.2\% to 85.8\% on out-of-distribution terrains. The student improves terrain traversal and contact quality while preserving flat-ground motion-tracking quality. On a Unitree G1, the student completes 117 of 120 trials with planned or VR-driven references, achieving 97.5\% success.
Project page: \url{https://perceptive-sonic.github.io/}.
\end{abstract}

% Overview float, placed after the abstract so it can appear on page 2.
\begin{figure*}[t]
    \centering
    \includegraphics[width=\textwidth]{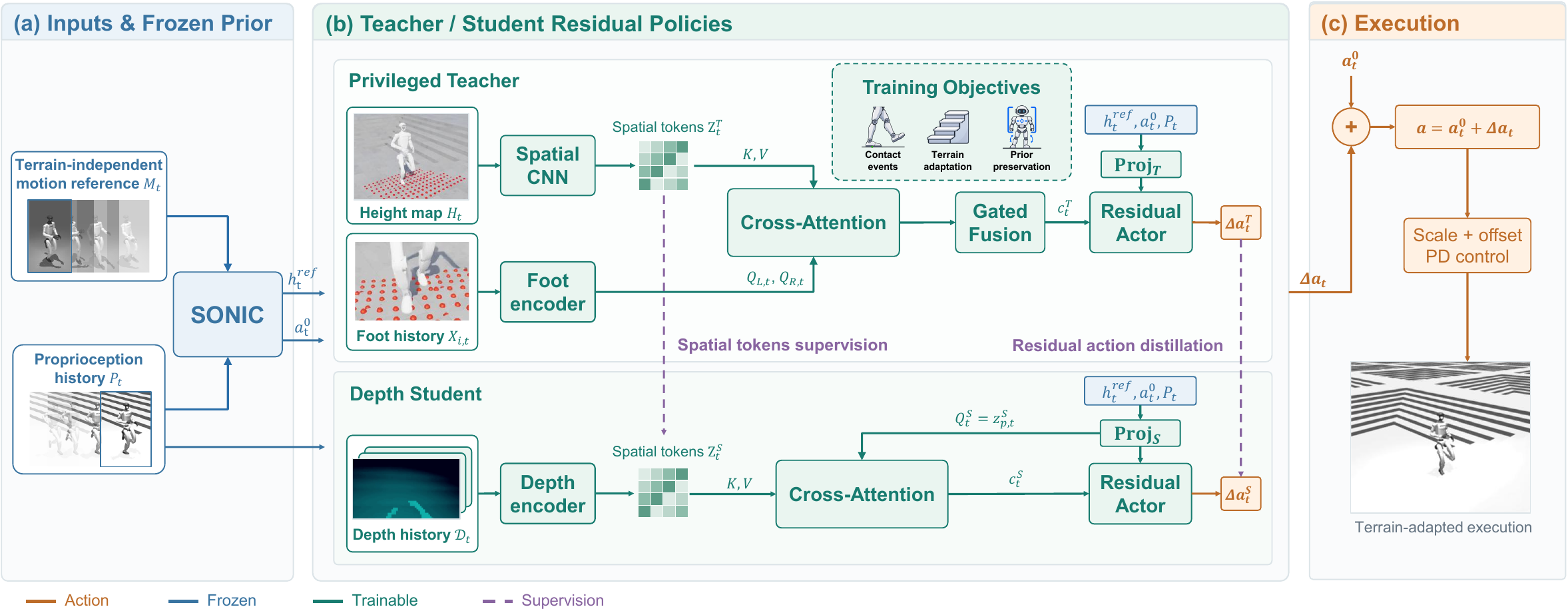}
    \caption{Perceptive SONIC. (a) Shared frozen SONIC prior. (b) Privileged teacher and depth student with independent feature projections. Teacher objectives combine motion tracking, reference-consistent contact, swept clearance, and residual regularization. Dashed links indicate online action and spatial-token distillation to the student. (c) The student residual augments the SONIC action for terrain-adaptive execution.}
    \label{fig:overview}
\end{figure*}

\section{Introduction}

Imitation learning and motion priors have enabled coordinated
humanoid behaviors~\cite{deepmimic,amp,phc}, with behavior
foundation models (BFMs) supporting diverse motion references
and control interfaces~\cite{exbody,exbody2,hover,beyondmimic,sonic}.
However, terrain-independent references may prescribe physically
infeasible contacts on stairs or obstacles. For example, a swing
trajectory that clears level ground can intersect the face of a step.
Perceptive locomotion uses maps or depth to anticipate
terrain~\cite{perceptive_wild,hpl,pim,hiking},
and recent work extends terrain awareness to motion
tracking~\cite{pgmt,perceptive_bfm}. Perceptive BFM synthesizes
paired raw and terrain-conformal references and trains a blind
teacher before student transfer~\cite{perceptive_bfm}.
The challenge remains how to endow pretrained BFMs with terrain
adaptability without relearning their motion priors or
compromising commanded behaviors.

We introduce \method{}, which decouples terrain adaptation
from pretrained whole-body coordination
(Figs.~\ref{fig:hardware_deployment} and~\ref{fig:overview}).
Our central insight is that terrain-aware execution can be
achieved by learning terrain-conditioned corrections to the
pretrained controller's actions rather than reconstructing
reference motions or relearning the entire tracking policy.
This retains the original motion reference as the command
interface throughout terrain adaptation. We therefore learn
joint-action residuals around frozen SONIC and regularize
flat-ground deviations and waist and arm corrections to
preserve motion fidelity as foot contacts adapt.

When reference motions conflict with the terrain, the first
challenge is to decide which aspects of contact to preserve.
We distinguish contact intent from its geometric realization.
Preserving a stance interval may still require shifting the
corresponding foothold onto a step. Foot-conditioned spatial
retrieval associates terrain geometry with each foot's support
requirements. Contact-event objectives
preserve reference-allowed stance intervals while permitting
foothold adjustment and discouraging mistimed, insufficient,
or repeated contacts. Swept clearance further anticipates
collision risks along upcoming reference-foot XY paths.
Together, these mechanisms preserve the intended contact
behavior while adapting its physical realization to terrain.

Even with these contact objectives, the residual policy must
infer the geometry needed to place and lift each foot from
partial, noisy onboard depth observations.
We therefore decouple terrain-aware control learning from
deployable perception learning through privileged
teacher--student distillation. Local height maps give the teacher
explicit terrain geometry during training. By jointly transferring
spatial terrain tokens and residual-action means on student-visited
states, the teacher supervises not only control decisions
but also their underlying geometric representations, enabling
near-teacher traversal performance from depth history
and proprioception.

To further evaluate \method{} in real-world environments,
we deploy the depth-based student on a physical Unitree G1
with motion references from SONIC Planner and VR teleoperation
(Fig.~\ref{fig:hardware_deployment}). The student succeeds
in 117 of 120 trials (97.5\%) and completes continuous runs of
30 minutes with SONIC Planner and 25 minutes with
VR teleoperation.

\section{Related Work}
\subsection{Humanoid Motion Tracking}
Human demonstrations specify either explicit tracking targets~\cite{deepmimic,phc,beyondmimic} or adversarial style rewards~\cite{amp}. For real humanoids, expressive controllers balance motion reproduction with stability~\cite{exbody,exbody2}, while teleoperation and multi-interface policies support different forms of human commands~\cite{h2o,omnih2o,hover}. SONIC scales this approach to general motion tracking~\cite{sonic}. Scene-associated data provide another route: VideoMimic reconstructs human--scene interactions, and GRAIL generates scene-conditioned demonstrations~\cite{videomimic,grail}. PGMT learns motion-conditioned terrain glimpses and relaxed tracking~\cite{pgmt}. Perceptive BFM synthesizes terrain-conformal references and trains an adapted-reference teacher before raw-reference distillation~\cite{perceptive_bfm}. Our framework retains the pretrained BFM unchanged and learns joint-action corrections through foot-conditioned terrain retrieval and contact rewards, without synthesizing terrain-adapted reference trajectories.

\subsection{Perceptive Legged Locomotion}
Large-scale simulation, history-based adaptation, and sim-to-real transfer methods support robust proprioceptive locomotion~\cite{walk_minutes,rma,humanoid_gym}. Exteroception enables anticipatory adjustment to terrain~\cite{perceptive_wild}, including depth-based parkour on quadrupeds and humanoids~\cite{extreme_parkour,robot_parkour,hpl}. Privileged terrain encoders and internal models further address noisy humanoid perception~\cite{pim,perceptive_humanoid}. Hiking in the Wild develops depth processing and strided history sampling~\cite{hiking}; SSR combines foresighted foothold guidance with symmetry and motion priors~\cite{ssr}. We study terrain adaptation under an externally specified whole-body motion reference, preserving that reference as the behavioral interface.

\subsection{Residual Adaptation and Privileged Transfer}
Residual reinforcement learning augments existing controllers with learned corrections~\cite{residual_rl,residual_policy}. ResMimic applies this principle to humanoid loco-manipulation~\cite{resmimic}. We use residual learning to adapt a frozen BFM to terrain, with foot-conditioned retrieval, reference-contact objectives, and prospective clearance guiding corrections toward reference-consistent support. Privileged learning exploits simulator state while restricting deployed policies to available observations~\cite{asymmetric,rma,exbody2}. To transfer this terrain-aware control to a depth-based student, we use online DAgger~\cite{pmlr-v15-ross11a} to jointly supervise residual actions and spatial tokens on student-visited states. The student learns control decisions and geometric features while retaining the frozen BFM, without privileged terrain maps or foot-state inputs.

\section{Method}
\label{sec:method}
To reconcile terrain-independent references with uneven support, \method{} learns residual corrections around a frozen BFM, reusing its whole-body coordination so training can focus on terrain adaptation~(Fig.~\ref{fig:overview}). A teacher--student framework separates learning terrain-aware control with privileged information from inferring corrections under restricted depth observations. During teacher training, contact and clearance rewards encourage terrain-compatible execution, while regularization losses preserve motion fidelity. The following subsections detail the residual architecture, teacher objectives, and depth-student distillation.

\subsection{Motion-Conditioned Residual Control}
We adopt a residual architecture both to preserve SONIC's pretrained tracking capability and to simplify policy learning by restricting it to the action corrections needed for terrain adaptation. For $n$ actuated degrees of freedom, frozen SONIC provides $a_t^0=\pi_0(M_t,P_t)\in\R^n$ from motion reference $M_t$ and proprioceptive history $P_t$, following its original interface~\cite{sonic}. The adaptor predicts
\begin{equation}
    \Delta a_t=\pi_\theta([z_{h,t},c_t,z_{a,t},z_{p,t}])\in\R^n,
    \label{eq:fusion}
\end{equation}
where $\Delta a_t$ is the residual-policy mean, conditioned on reference motion, terrain geometry, base action, and proprioception. During teacher training, $a_t\sim\mathcal N(a_t^0+\Delta a_t,\operatorname{diag}(\sigma^2))$, where $\sigma$ contains learnable, state-independent joint standard deviations. At inference, the mean action is
\begin{equation}
    a_t=a_t^0+\Delta a_t,
    \label{eq:action}
\end{equation}
followed by proportional--derivative (PD) control~(Fig.~\ref{fig:overview}(c)).

Encoders give $z_{h,t}=E_h(h_t^{\mathrm{ref}})$, $z_{a,t}=E_a(a_t^0)$, and $z_{p,t}=E_p(P_t)$, where $h_t^{\mathrm{ref}}$ is SONIC's pre-quantization reference feature. Each feature and context $c_t$ has 256 channels. Superscripts $T/S$ denote teacher/student. Their projections $\operatorname{Proj}_T,\operatorname{Proj}_S$ and residual actors are independent, sharing frozen SONIC~(Fig.~\ref{fig:overview}(b)).

\subsection{Privileged Terrain Teacher and Objectives}
\label{sec:teacher}
\label{sec:terrain_objective}
Exact foot tracking can cause collisions or poor support. To adapt footholds and clearance while preserving contact intent, the teacher uses privileged terrain and foot states~(Fig.~\ref{fig:overview}(b)). We formulate a terrain-conditioned partially observable Markov decision process (POMDP)
\begin{equation}
    (\mathcal S,\mathcal O_T,\mathcal A_\Delta,
      \mathcal P_\xi,r,\gamma).
    \label{eq:teacher_pomdp}
\end{equation}
The state space $\mathcal S$ includes robot and simulator state, the motion reference and phase, and observation/contact-event histories. The teacher observes $o_t^T=(M_t,P_t,H_t,X_{L,t},X_{R,t})\in\mathcal O_T$, with local height map $H_t$ and foot-state histories $X_{i,t}$. Residual actions belong to $\mathcal A_\Delta=\R^n$. Kernel $\mathcal P_\xi$ describes dynamics on terrain $\xi$ under $a_t^0+\Delta a_t^T$. The normalized curriculum level is $\ell_t\in[0,1]$, and $\gamma$ is the discount factor. We train with proximal policy optimization (PPO)~\cite{ppo}. The reward combines three components:
\begin{equation}
    r=r_{\mathrm{track}}+r_{\mathrm{adapt}}+r_{\mathrm{reg}}.
    \label{eq:teacher_reward}
\end{equation}
Here $r_{\mathrm{track}}$ rewards motion tracking, $r_{\mathrm{adapt}}$ promotes terrain traversal and reference-consistent contact, and $r_{\mathrm{reg}}$ regularizes motion and penalizes unsafe behavior.

\textit{Foot-conditioned terrain retrieval.}
To better retrieve terrain information needed for foot motion, we use foot-conditioned queries. A spatial encoder maps torso-yaw-aligned $H_t\in\R^{16\times16}$ to tokens $Z_t^T=E_H(H_t)\in\R^{16\times64}$ with scanner-coordinate positional encoding. To query these tokens for foot $i\in\{L,R\}$, we construct $X_{i,t}$ from ten-step XY positions normalized in the torso-yaw scan frame, together with the root's planar velocity and yaw rate expressed in the root body frame. A shared encoder with learned side embedding $v_i$ forms $Q_{i,t}=E_X(X_{i,t},v_i)$ to retrieve features from $Z_t^T$ through cross-attention~\cite{attention}. The retrieved features are then refined by query-residual addition and layer normalization before softmax-gated pooling and projection yield $c_t^T$.

\textit{Difficulty-adaptive tracking.}
To accommodate terrain-driven leg corrections, $r_{\mathrm{track}}$ relaxes lower-body tracking while retaining strict upper-body tracking. Position and orientation rewards apply Gaussian kernels to $[e_{\mathcal L}^p-\delta_p(\ell_t)]_+$ and $[e_{\mathcal L}^R-\delta_R(\ell_t)]_+$, where $[x]_+=\max(0,x)$. The errors compare lower-body link positions and orientations with the reference in meters and radians. The tolerances $\delta_p(\ell_t)=0.02+0.08\ell_t$\,m and $\delta_R(\ell_t)=0.10+0.20\ell_t$\,rad increase with terrain difficulty, allowing larger deviations before reward decay.

\begin{figure}[t]
    \centering
    \includegraphics[width=0.90\columnwidth]{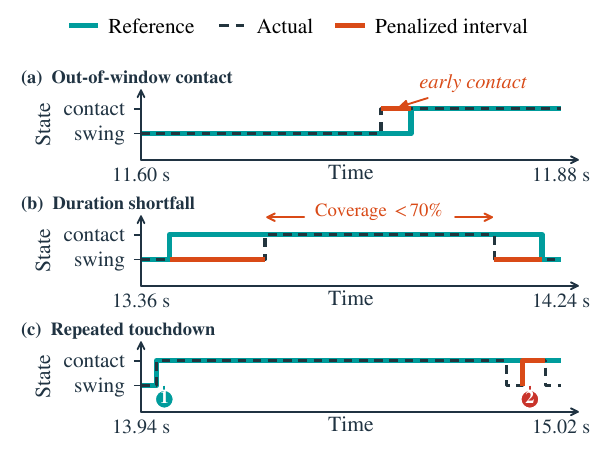}
    \caption{Reference-contact event penalties:
    (a) out-of-window contact;
    (b) insufficient stance coverage ($<70\%$);
    (c) repeated touchdown.
    Teal: reference; dashed black: actual; orange: penalized intervals.}
    \label{fig:contact_event_penalties}
\end{figure}

\textit{Reference-consistent contact.}
The support term $r_{\mathrm{support}}$ in $r_{\mathrm{adapt}}$ rewards each foot's supported sole-area fraction during reference-allowed contact intervals. For each foot $i$, $\rho_{i,t}$ labels reference stance, $\kappa_{i,t}$ gives its confidence, $\alpha_{i,t}$ indicates actual contact, and $\eta_{i,t}$ measures sole support:
\begin{equation}
    r_{\mathrm{support}}=
    \frac{\sum_{i\in\{L,R\}}\kappa_{i,t}\rho_{i,t}\alpha_{i,t}\eta_{i,t}}
         {\sum_{i\in\{L,R\}}\kappa_{i,t}\rho_{i,t}},
    \label{eq:legal_support}
\end{equation}
$r_{\mathrm{support}}$ is zero when its denominator is zero. Three confidence-weighted contact-event penalties discourage reference-flight contact, stance coverage below 70\%, and repeated touchdowns within a reference interval~(Fig.~\ref{fig:contact_event_penalties}).

\begin{figure}[t]
    \centering
    \includegraphics[width=0.855\columnwidth]{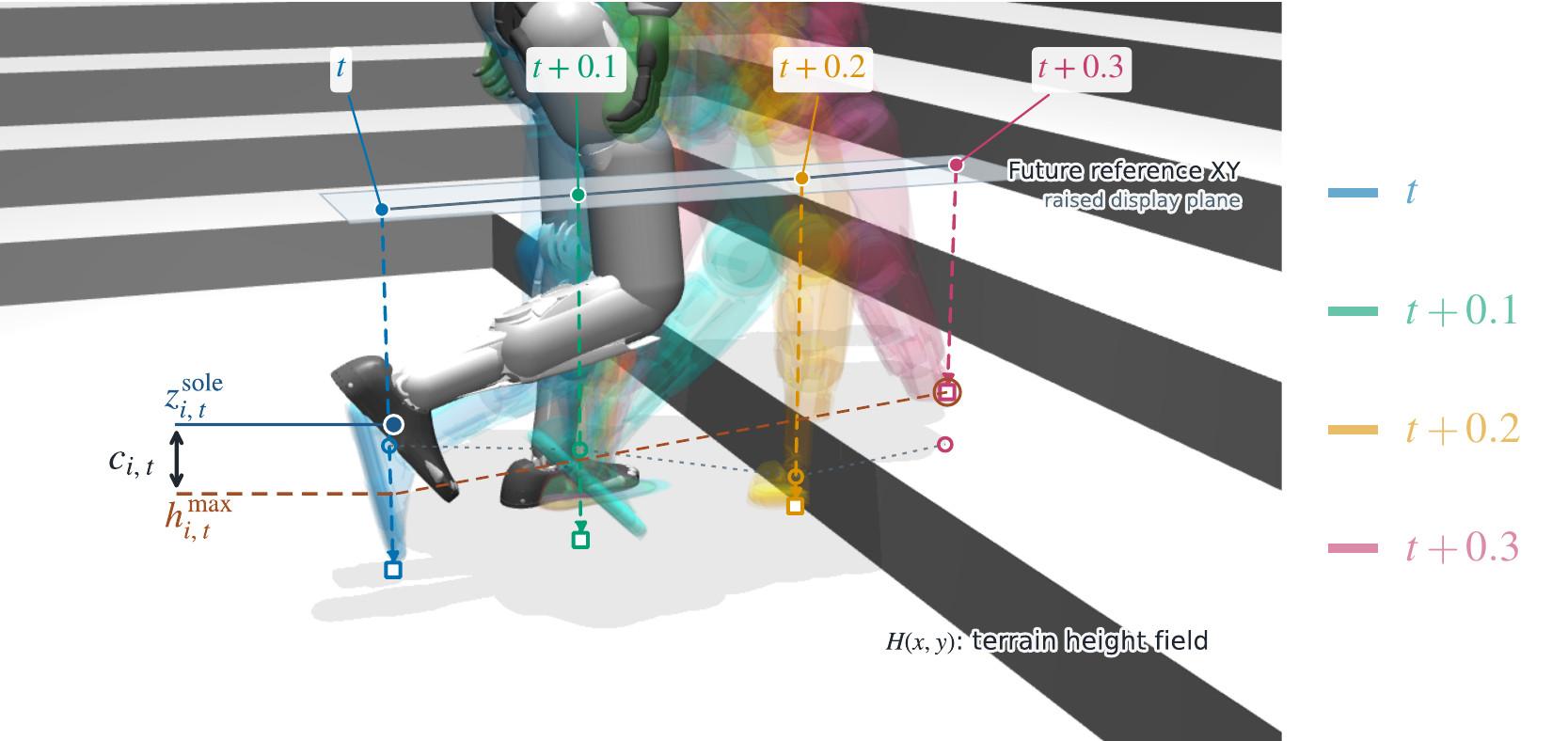}
    \caption{Swept-clearance reward along re-anchored reference-foot XY locations. Hollow circles show reference feet; raised markers indicate XY queries. }
    \label{fig:swept_clearance}
\end{figure}

\textit{Prospective swept clearance.}
The swept-clearance term in $r_{\mathrm{adapt}}$ encourages anticipatory foot lifting along upcoming reference paths~(Fig.~\ref{fig:swept_clearance}). Let $x_t,x_t^{\mathrm{ref}}$ and $\psi_t,\psi_t^{\mathrm{ref}}$ be robot/reference root XY positions and yaws. Re-anchor future reference-foot XY locations by
\begin{equation}
    \widetilde p_{i,t+\tau_k}^{\mathrm{ref},xy}
    =x_t+R(\psi_t-\psi_t^{\mathrm{ref}})
      (p_{i,t+\tau_k}^{\mathrm{ref},xy}-x_t^{\mathrm{ref}}),
    \label{eq:reference_reanchor}
\end{equation}
where $R$ is planar rotation and $\tau_k=0.1k$\,s for $k\in\{0,1,2,3\}$. Privileged terrain query $\mathcal H$ gives
$h_{i,t}^{\max}=\max_k\mathcal H(\widetilde p_{i,t+\tau_k}^{\mathrm{ref},xy})$.
\begin{samepage}
With current sole height $z_{i,t}^{\mathrm{sole}}$ expressed in the same vertical frame as $\mathcal H$, clearance $c_{i,t}=z_{i,t}^{\mathrm{sole}}-h_{i,t}^{\max}$ determines
\begin{equation}
    r_{\mathrm{swept}}=\frac12\sum_{i\in\{L,R\}}b_{i,t}^{\mathrm{sw}}
        \exp\!\left[-\left(\frac{[d_{\mathrm{clr}}-c_{i,t}]_+}{\sigma_{\mathrm{clr}}}\right)^2\right].
    \label{eq:swept}
\end{equation}
\end{samepage}
Target $d_{\mathrm{clr}}$ and scale $\sigma_{\mathrm{clr}}$ shape the reward. Gate $b_{i,t}^{\mathrm{sw}}$ uses actual low-force swing status independently of reference swing labels.

\textit{Residual preservation and optimization.}
To preserve SONIC's flat-ground tracking and upper-body motion during terrain adaptation, we use flat-identity (FI) and upper-body/waist (UW) residual losses
\begin{equation}
\begin{aligned}
    \mathcal L_{\mathrm{FI}}&=\mathbb{E}_t\!\left[m_t\|\Delta a_t^T\|_2^2\right],\\
    \mathcal L_{\mathrm{UW}}&=\mathbb{E}_t\!\left[|\mathcal J|^{-1}
       \sum_{j\in\mathcal J}(\Delta a_{t,j}^T)^2\right],
\end{aligned}
\label{eq:residual_preservation}
\end{equation}
where $m_t$ indicates flat terrain from height-map variance, $\mathcal J$ contains waist and arm joints, and expectations average over rollout states. The teacher loss is
\begin{equation}
    \mathcal L_T=\mathcal L_{\mathrm{PPO}}
    +\lambda_{\mathrm{FI}}\mathcal L_{\mathrm{FI}}
    +\lambda_{\mathrm{UW}}\mathcal L_{\mathrm{UW}}.
    \label{eq:teacher_loss}
\end{equation}
Here $\lambda_{\mathrm{FI}}$ and $\lambda_{\mathrm{UW}}$ weight the auxiliary losses, and $\mathcal L_{\mathrm{PPO}}$ includes policy, value, and entropy terms.

\subsection{Depth Student and Online Distillation}
\label{sec:student}
Learning control from depth couples visual inference with complex control exploration. We distill the privileged teacher's residual actions and spatial tokens to separate these tasks, using local height-map features to supervise multi-frame depth fusion.

Each depth frame is preprocessed as in~\cite{hiking} to obtain an $18\times32$ image. Ten frames sampled at a five-frame stride form $\mathcal D_t\in[0,1]^{10\times18\times32}$. An encoder with learned positional embeddings produces $Z_t^S=E_D(\mathcal D_t)\in\R^{16\times64}$. Projected tokens supply attention keys/values and $Q_t^S=z_{p,t}^S$ supplies the query. Cross-attention with query-residual addition and layer normalization gives $c_t^S$. The residual actor also receives $z_{p,t}^S$ directly~\eqref{eq:fusion}.

Online DAgger-style distillation~\cite{pmlr-v15-ross11a} supervises student-visited states. The student executes $\Delta a_t^S$, while the frozen teacher supplies residual-action means $\Delta a_t^T$ and pre-attention spatial tokens, following the dashed supervision links in Fig.~\ref{fig:overview}(b):
\begin{equation}
\begin{aligned}
    \mathcal L_D={}&\operatorname{Huber}_1(\Delta a_t^S,\sg[\Delta a_t^T])\\
      &+0.5\operatorname{Huber}_1(Z_t^S,\sg[Z_t^T]).
\end{aligned}
    \label{eq:distill}
\end{equation}
Here $\sg$ stops teacher gradients. Both unit-threshold Huber terms average over samples and features.

\section{Experiments}
\label{sec:experiments}
We evaluate traversal and motion quality, assess teacher--student transfer, and examine the contribution of each design choice.

% Generated by scripts/prepare_evidence.py; do not edit numbers.
\begin{table*}[t]
\centering
\caption{Full-clip simulation performance. Touchdown (TD) support measures landings with at least 50\% sole support; allowed TD measures landings by a reference-permitted foot. Slip is mean horizontal stance-foot speed.}
\label{tab:main}
\small
\setlength{\tabcolsep}{4pt}
\begin{tabular*}{\textwidth}{@{\extracolsep{\fill}}lrrrrrr@{}}
\toprule
Policy & \shortstack{Success\\(\%) $\uparrow$} & \shortstack{Hard fail\\(\%) $\downarrow$} & \shortstack{TD support\\(\%) $\uparrow$} & \shortstack{Allowed TD\\(\%) $\uparrow$} & \shortstack{Slip\\(m/s) $\downarrow$} & \shortstack{MPJPE-L\\(mm) $\downarrow$} \\
\midrule
SONIC & 58.055 & 40.298 & 59.98 & 63.93 & 0.09887 & 36.267 \\
Privileged teacher (ours) & 91.526 & 4.601 & \textbf{81.97} & \textbf{70.88} & 0.05007 & 36.352 \\
Depth student (ours) & \textbf{91.623} & \textbf{4.548} & \textbf{81.97} & 70.40 & \textbf{0.04984} & 35.933 \\
Blind distilled student & 90.216 & 5.449 & 80.95 & 67.74 & 0.05013 & \textbf{35.753} \\
Teacher w/o swept clearance & 90.982 & 4.663 & 80.44 & 60.97 & 0.05097 & 36.140 \\
\bottomrule
\end{tabular*}
\end{table*}

\subsection{Setup and Evaluation Protocol}
\label{sec:protocol}
We use a 29-joint Unitree G1 with $200$\,Hz physics and $50$\,Hz control. The training set contains 18,562 clips (44.24\,h) from BONES-SEED~\cite{bones_seed}, MotionDecode~\cite{motiondecode}, and AMASS~\cite{AMASS:ICCV:2019}. We select foot-supported bipedal motions with travel aligned to root yaw, using SONIC robot motion commands~\cite{sonic}. Contacts are annotated at 50\,Hz from sole-point stationarity~\cite{scenebot}. Stance cores require speed below $0.12$\,m/s, acceleration below $3$\,m/s$^2$, and a $2.5$\,cm stationarity radius within $0.1$\,s. Hysteresis uses a $0.22$\,m/s release threshold and $0.06$\,s confirmation.

Both stages train in IsaacLab/IsaacSim~\cite{isaaclab} on two NVIDIA RTX 4090 GPUs, with $\lambda_{\mathrm{FI}}=0.03$ and $\lambda_{\mathrm{UW}}=0.01$. Training terrains occupy $10\times10$\,m fields with flat ground, regular equal-height stairs, rough ground, slopes, random grids, and discrete obstacles~(Fig.~\ref{fig:all_terrains}(a)--(g)). Architecture, optimization, and rewards appear in the Appendix.

\begin{figure}[t]
    \centering
    \includegraphics[width=\columnwidth]{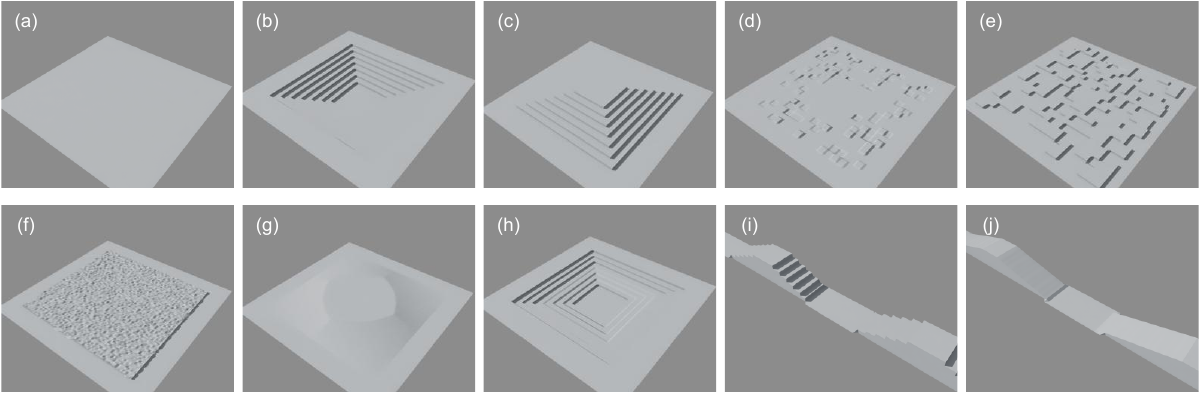}
    \caption{Training terrains comprise (a) flat ground, (b) inverse pyramid stairs, (c) pyramid stairs, (d) discrete obstacles, (e) random grid, (f) rough ground, and (g) slope. OOD terrains comprise (h) random-height stairs, (i) long stairs, and (j) long slope.}
    \label{fig:all_terrains}
\end{figure}

From 7,889 held-out motions, we exclude AMASS clips for arm-induced camera occlusion and select 2,048 using category-proportional quotas allocated by the largest remainder method. Evaluation covers 20,758 full-clip episodes across these motions (episode-weighted median duration $5.20$\,s, range $1.97$--$122.87$\,s). Success requires no hard termination, trajectory-tracking progress above $0.75$, and no stall exceeding $1$\,s.

To test terrain generalization, we execute test motions on OOD terrains in 6,422 episodes~(Fig.~\ref{fig:all_terrains}(h)--(j)). Independently randomized stair heights test local geometric irregularity. Long stairs and slopes use $18\times1.2$\,m channels with continuous segments and platforms, testing sustained traversal, narrow support, and lateral tracking. These courses vary length, layout, and width relative to the training pyramid stairs and slopes. References are filtered to fit the narrow courses.

\subsection{Residual Control and Motion Quality}
\label{sec:sonic_baseline}
We compare \method{} with SONIC on full-clip traversal, contact quality, and flat-ground tracking to test whether terrain adaptation preserves motion quality.

\begin{figure}[t]
    \centering
    \includegraphics[width=\columnwidth]{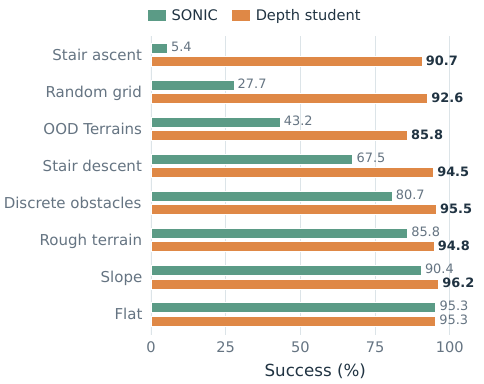}
    \caption{Traversal success of SONIC and the depth student across terrains.}
    \label{fig:terrain_success_all}
\end{figure}

\textit{Terrain traversal.} The depth student raises overall full-clip success from $58.05\%$ to $91.62\%$~(Table~\ref{tab:main}), and OOD success from $43.21\%$ to $85.80\%$~(Fig.~\ref{fig:terrain_success_all}).

\textit{Support and gait consistency.} To assess support quality, we measure touchdown (TD) support (\%), the fraction of landings with at least 50\% sole support, and slip, the mean stance-foot horizontal speed. Allowed TD reports landings by reference-permitted feet. Timing and Extra TD assess gait consistency and movement stability. Timing measures the mean absolute error (MAE) between matched same-foot reference and actual onsets. Extra TD counts additional touchdowns per observed reference interval (\%), while fragmentation measures intervals with broken contact. The teacher approximately halves slip, improves touchdown support, and reduces contact interruptions~(Tables~\ref{tab:main} and~\ref{tab:contact_quality}), indicating more stable, reference-consistent support.

% Generated by scripts/prepare_evidence.py; do not edit numbers.
\begin{table}[t]
\centering
\caption{Reference-contact quality. Timing: onset MAE; Extra TD: additional touchdowns divided by observed reference intervals, reported in percent; Frag.: fragmented intervals.}
\label{tab:contact_quality}
\small
\setlength{\tabcolsep}{2.5pt}
\renewcommand{\arraystretch}{1.15}
\begin{tabular*}{\columnwidth}{@{\extracolsep{\fill}}lrrr@{}}
\toprule
Policy & \shortstack{Timing\\(ms) $\downarrow$} & \shortstack{Extra TD\\(\%) $\downarrow$} & \shortstack{Frag.\\(\%) $\downarrow$} \\
\midrule
SONIC & 46.35 & 17.86 & 13.11 \\
\addlinespace[2pt]
Privileged teacher (ours) & 44.75 & 9.95 & 7.45 \\
\addlinespace[2pt]
Depth student (ours) & \textbf{44.23} & \textbf{9.87} & \textbf{7.40} \\
\addlinespace[2pt]
Teacher w/o cross-attention & 45.32 & 11.58 & 8.39 \\
\addlinespace[2pt]
Teacher w/o contact penalties & 46.39 & 11.18 & 8.20 \\
\bottomrule
\end{tabular*}
\end{table}

\textit{Motion preservation.} Flat-ground evaluation tests whether adaptation retains the original BFM's tracking quality~(Table~\ref{tab:flat}). Local mean per-joint position error (MPJPE-L) measures overall motion-tracking quality after removing pelvis translation while retaining yaw. The privileged teacher and depth student retain tracking accuracy close to SONIC. The student's flat-ground success remains nearly unchanged, while no-fall survival shows an absolute increase of $0.44\%$ and stance slip decreases by $26.24\%$.

% Generated by scripts/prepare_evidence.py; do not edit numbers.
\begin{table}[t]
\centering
\caption{Traversal and motion preservation. Overall success covers all terrains. Flat metrics use flat ground.}
\label{tab:flat}
\small
\setlength{\tabcolsep}{2pt}
\begin{tabular*}{\columnwidth}{@{\extracolsep{\fill}}lrrr@{}}
\toprule
Policy & \shortstack{Overall\\success\\(\%) $\uparrow$} & \shortstack{Flat\\success\\(\%) $\uparrow$} & \shortstack{Flat\\MPJPE-L\\(mm) $\downarrow$} \\
\midrule
SONIC & 58.05 & 95.31 & \textbf{26.86} \\
Privileged teacher (ours) & 91.53 & 96.00 & 27.01 \\
Depth student (ours) & 91.62 & 95.26 & 26.96 \\
Direct decoder FT & \textbf{92.85} & \textbf{96.19} & 30.01 \\
\bottomrule
\end{tabular*}
\end{table}

Together, these results show that residual adaptation improves terrain traversal and contact quality while retaining flat-ground motion-tracking quality.

\subsection{Teacher--Student Transfer}
We assess whether distillation preserves privileged control by comparing teacher and student. Their absolute success gap is within $0.10\%$, with comparable tracking, support, and contact quality~(Tables~\ref{tab:main} and~\ref{tab:contact_quality}), demonstrating near-teacher performance from depth history and proprioception.

\subsection{Ablation Studies}
Ablations examine residual control, perception, and teacher objectives.

\begin{figure}[t]
    \centering
    \includegraphics[width=\columnwidth]{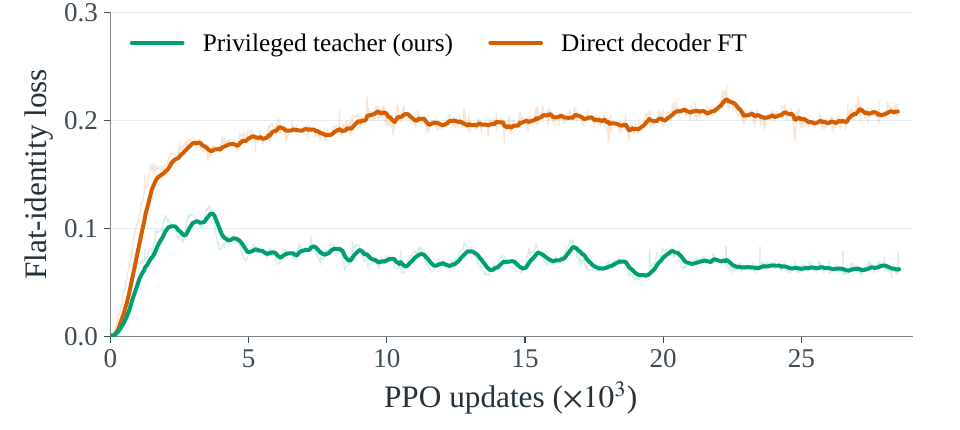}
    \caption{Flat-identity loss with FI/UW regularization. Faint curves show spike-filtered samples; solid curves show smoothed trends.}
    \label{fig:flat_identity_loss}
\end{figure}

\textit{Direct decoder fine-tuning (FT).} To assess whether freezing preserves motion quality, we replace the residual actor with a trainable SONIC decoder, retaining the terrain encoder and foot-query attention. Its final hidden layer receives zero-initialized terrain modulation, while the motion encoder and quantizer remain frozen. FT action deviations are measured against frozen SONIC at matched states. Fine-tuning incurs higher flat-identity loss~(Fig.~\ref{fig:flat_identity_loss}). It raises overall success from $91.53\%$ to $92.85\%$ but increases flat-ground MPJPE-L from $27.01$ to $30.01$\,mm~(Table~\ref{tab:flat}). Frozen residual adaptation thus retains flat-ground tracking accuracy closer to SONIC.

\textit{Visual input.} We assess the contribution of depth by distilling a blind student from the same privileged teacher. Its BFM and distillation objectives match those of the depth student, with depth inputs set to zero throughout training. Without vision, overall success shows an absolute drop of $1.41\%$, and the collision-free success proxy (success without detected kicks or non-foot contacts) falls from $28.06\%$ to $22.54\%$. Allowed TD also shows an absolute drop of $2.66\%$~(Table~\ref{tab:main}), indicating poorer reference-contact consistency and less stable contact.

\textit{Cross-attention.} To assess foot-conditioned spatial retrieval, we replace cross-attention with a shared MLP that fuses terrain tokens with each foot query. Without cross-attention, the extra-touchdown rate is $16.33\%$ higher than that of the privileged teacher and $17.35\%$ higher than that of the depth student~(Table~\ref{tab:contact_quality}). Fragmentation also increases, with larger extra-touchdown increases on long stairs and slopes. Cross-attention also raises OOD success from $83.87\%$ to $85.43\%$, with an absolute overall success gain of $0.80\%$. These results support spatial retrieval for traversal and contact quality.

\textit{Swept clearance.} Removing prospective clearance supervision tests whether upcoming terrain heights help guide foot clearance. This reduces allowed TD from $70.88\%$ to $60.97\%$, an absolute drop of $9.91\%$, alongside lower touchdown support and traversal success~(Table~\ref{tab:main}). The swept-clearance reward improves both traversal success and consistency with reference-allowed foot contacts.

\textit{Contact-event penalties.} We train a teacher without the three contact-event penalties in Fig.~\ref{fig:contact_event_penalties} to assess contact interruptions and extra touchdowns. Fragmentation rises from $7.45\%$ to $8.20\%$, with more extra touchdowns~(Table~\ref{tab:contact_quality}), supporting their use for reference-consistent contact.

\subsection{Real-Robot and Qualitative Evaluation}
\label{sec:hardware}
We test deployment across motion interfaces on Unitree G1 over flat ground, stairs in both directions, grass, and random obstacles. References come from gamepad-operated SONIC Planner or VR teleoperation~(Fig.~\ref{fig:hardware_deployment}). Hardware success requires motion completion without a fall, while survival measures the absence of falls.

Across 120 trials, the student achieves 117 successes (97.5\%) and 119 survivals (99.2\%)~(Table~\ref{tab:hardware}). All 60 Planner-driven trials succeed, alongside 57 of 60 VR trials, supporting both control interfaces across the tested terrains.

Long-duration tests assess sustained execution. SONIC Planner completes an uninterrupted 30-minute outdoor run with 14 stair traversals, including ascent and descent, and level-ground walking, without resets, manual stabilization, or falls. VR teleoperation completes a continuous 25-minute run with 10 stair traversals and level-ground walking without falls, with the operator on level ground throughout.

% Generated by scripts/prepare_hardware_table.py.
\begin{table}[t]
\centering
\caption{Real-world deployment results (trial counts).}
\label{tab:hardware}
\small
\setlength{\tabcolsep}{3pt}
\begin{tabular}{llrrr}
\toprule
Terrain & Motion input & Trials & Success & Survival \\
\midrule
Flat & SONIC Planner & 10 & 10 & 10 \\
Flat & VR teleop. & 10 & 10 & 10 \\
Stairs up & SONIC Planner & 15 & 15 & 15 \\
Stairs up & VR teleop. & 15 & 14 & 15 \\
Stairs down & SONIC Planner & 15 & 15 & 15 \\
Stairs down & VR teleop. & 15 & 14 & 14 \\
Random obstacles & SONIC Planner & 20 & 20 & 20 \\
Random obstacles & VR teleop. & 20 & 19 & 20 \\
\midrule
Total & --- & 120 & 117 & 119 \\
\bottomrule
\end{tabular}
\end{table}

\section{Conclusion}
\method{} adapts a frozen BFM through joint-action residuals guided by foot-conditioned terrain retrieval, reference-contact supervision, and swept clearance. Online distillation transfers these corrections to a depth student. Simulation and real-robot results show improved traversal and contact quality without terrain-specific demonstrations. Future work will extend adaptation to whole-body contacts and enable zero-shot substitution of the base BFM.

\section*{Acknowledgment}
The authors acknowledge the support of OPPO Research Institute in providing robotic platforms and computational resources for this research.
% Activate balancing within the first column of the final References page.
\IEEEtriggeratref{20}
\IEEEtriggercmd{\flushcolsend}
\bibliographystyle{IEEEtran}
\bibliography{references}

@misc{scenebot,
  author = {Sirui Chen and Shibo Zhao and Zhen Wu and Jiaman Li and Guanya Shi and C. Karen Liu},
  title = {{SceneBot}: Contact-Prompted General Humanoid Whole Body Tracking with Scene-Interaction},
  year = {2026},
  howpublished = {arXiv preprint arXiv:2606.27581},
  eprint = {2606.27581},
  archiveprefix = {arXiv},
  primaryclass = {cs.RO},
  url = {https://arxiv.org/abs/2606.27581},
}

@article{sonic,
  author = {Luo, Zhengyi and Yuan, Ye and Wang, Tingwu and Li, Chenran and Castañeda, Fernando and Chen, Sirui and Cao, Zi-Ang and Li, Jiefeng and Minor, David and Ben, Qingwei and Park, Jinhyung and Sami, David and Wang, Zi and Da, Xingye and Ding, Runyu and Hogg, Cyrus and Song, Lina and Lim, Edy and Jeong, Eugene and He, Tairan and Xue, Haoru and Xiao, Wenli and Yuen, Simon and Kautz, Jan and Chang, Yan and Iqbal, Umar and Fan, Linxi Jim and Zhu, Yuke},
  title = {{SONIC}: Supersizing motion tracking for natural humanoid whole-body control},
  journal = {Sci. Robot.},
  volume = {11},
  number = {117},
  articleno = {eaed4592},
  year = {2026},
  note = {{Art. no.} eaed4592},
}

@misc{ppo,
  author = {John Schulman and Filip Wolski and Prafulla Dhariwal and Alec Radford and Oleg Klimov},
  title = {Proximal Policy Optimization Algorithms},
  year = {2017},
  howpublished = {arXiv preprint arXiv:1707.06347},
  eprint = {1707.06347},
  archiveprefix = {arXiv},
  primaryclass = {cs.LG},
  url = {https://arxiv.org/abs/1707.06347},
}

@article{deepmimic,
  author = {Peng, Xue Bin and Abbeel, Pieter and Levine, Sergey and van de Panne, Michiel},
  title = {{DeepMimic}: Example-guided Deep Reinforcement Learning of Physics-based Character Skills},
  journal = {ACM Trans. Graph.},
  volume = {37},
  number = {4},
  articleno = {143},
  numpages = {14},
  year = {2018},
  note = {{Art. no.} 143},
}

@article{amp,
  author = {Peng, Xue Bin and Ma, Ze and Abbeel, Pieter and Levine, Sergey and Kanazawa, Angjoo},
  title = {{AMP}: Adversarial Motion Priors for Stylized Physics-Based Character Control},
  journal = {ACM Trans. Graph.},
  volume = {40},
  number = {4},
  articleno = {144},
  numpages = {20},
  year = {2021},
  note = {{Art. no.} 144},
}

@inproceedings{phc,
  author = {Luo, Zhengyi and Cao, Jinkun and Winkler, Alexander and Kitani, Kris and Xu, Weipeng},
  title = {Perpetual Humanoid Control for Real-time Simulated Avatars},
  booktitle = {Proc. IEEE/CVF Int. Conf. Comput. Vis. (ICCV)},
  pages = {10861--10870},
  year = {2023},
}

@inproceedings{h2o,
  author = {He, Tairan and Luo, Zhengyi and Xiao, Wenli and Zhang, Chong and Kitani, Kris and Liu, Changliu and Shi, Guanya},
  title = {Learning Human-to-Humanoid Real-Time Whole-Body Teleoperation},
  booktitle = {Proc. IEEE/RSJ Int. Conf. Intell. Robots Syst. (IROS)},
  pages = {8944--8951},
  year = {2024},
}

@inproceedings{omnih2o,
  author = {He, Tairan and Luo, Zhengyi and He, Xialin and Xiao, Wenli and Zhang, Chong and Zhang, Weinan and Kitani, Kris M. and Liu, Changliu and Shi, Guanya},
  title = {{OmniH2O}: Universal and Dexterous Human-to-Humanoid Whole-Body Teleoperation and Learning},
  booktitle = {Proc. 8th Conf. Robot Learn. (CoRL)},
  series = {Proceedings of Machine Learning Research},
  volume = {270},
  pages = {1516--1540},
  year = {2025},
  url = {https://proceedings.mlr.press/v270/he25b.html},
}

@inproceedings{exbody,
  author = {Xuxin Cheng AND Yandong Ji AND Junming Chen AND Ruihan Yang AND Ge Yang AND Xiaolong Wang},
  title = {Expressive Whole-Body Control for Humanoid Robots},
  booktitle = {Proc. Robot. Sci. Syst. (RSS)},
  year = {2024},
}

@inproceedings{exbody2,
  author = {Ji, Mazeyu and Peng, Xuanbin and Liu, Fangchen and Li, Jialong and Yang, Ge and Cheng, Xuxin and Wang, Xiaolong},
  title = {{ExBody2}: Advanced Expressive Humanoid Whole-Body Control},
  booktitle = {Proc. IEEE Int. Conf. Robot. Autom. (ICRA)},
  pages = {613--620},
  year = {2026},
}

@inproceedings{hover,
  author = {He, Tairan and Xiao, Wenli and Lin, Toru and Luo, Zhengyi and Xu, Zhenjia and Jiang, Zhenyu and Kautz, Jan and Liu, Changliu and Shi, Guanya and Wang, Xiaolong and Fan, Linxi Jim and Zhu, Yuke},
  title = {{HOVER}: Versatile Neural Whole-Body Controller for Humanoid Robots},
  booktitle = {Proc. IEEE Int. Conf. Robot. Autom. (ICRA)},
  pages = {9989--9996},
  year = {2025},
}

@article{beyondmimic,
  author = {Liao, Qiayuan and Truong, Takara E. and Huang, Xiaoyu and Gao, Yuman and Tevet, Guy and Sreenath, Koushil and Liu, C. Karen},
  title = {{BeyondMimic}: From motion tracking to versatile humanoid control via guided diffusion},
  journal = {Sci. Robot.},
  volume = {11},
  number = {117},
  articleno = {eadx8924},
  year = {2026},
  note = {{Art. no.} eadx8924},
}

@inproceedings{resmimic,
  author = {Siheng Zhao and Yanjie Ze and Yue Wang and C. Karen Liu and Pieter Abbeel and Guanya Shi and Rocky Duan},
  title = {{ResMimic}: From General Motion Tracking to Humanoid Whole-body Loco-Manipulation via Residual Learning},
  booktitle = {Proc. IEEE-RAS Int. Conf. Humanoid Robots (Humanoids)},
  year = {2026},
  eprint = {2510.05070},
  archiveprefix = {arXiv},
  primaryclass = {cs.RO},
  url = {https://arxiv.org/abs/2510.05070},
  note = {to appear},
}

@inproceedings{grail,
  author = {Tianyi Xie and Haotian Zhang and Jinhyung Park and Zi Wang and Bowen Wen and Jiefeng Li and Xueting Li and Qingwei Ben and Haoyang Weng and Yufei Ye and David Minor and Tingwu Wang and Chenfanfu Jiang and Sanja Fidler and Jan Kautz and Linxi Fan and Yuke Zhu and Zhengyi Luo and Umar Iqbal and Ye Yuan},
  title = {{GRAIL}: Generating Humanoid Loco-Manipulation from World Models},
  booktitle = {Proc. 10th Conf. Robot Learn. (CoRL)},
  year = {2026},
  url = {https://arxiv.org/abs/2606.05160},
}

@inproceedings{hiking,
  author = {Zhu, Shaoting and Zhuang, Ziwen and Zhao, Mengjie and Lee, Kun-Ying and Zhao, Hang},
  title = {Hiking in the Wild: A Scalable Perceptive Parkour Framework for Humanoids},
  booktitle = {Proc. 10th Conf. Robot Learn. (CoRL)},
  year = {2026},
  url = {https://arxiv.org/abs/2601.07718},
}

@inproceedings{ssr,
  author = {Yu, Ruiqi and Wang, Yiwen and Hao, Yuan and Wu, Jun and Zhu, Qiuguo},
  title = {{SSR}: Scaling Surefooted and Symmetric Humanoid Traversal Toward the Open World},
  booktitle = {Proc. 10th Conf. Robot Learn. (CoRL)},
  year = {2026},
  url = {https://arxiv.org/abs/2605.30770},
}

@inproceedings{hpl,
  author = {Zhuang, Ziwen and Yao, Shenzhe and Zhao, Hang},
  title = {Humanoid Parkour Learning},
  booktitle = {Proc. 8th Conf. Robot Learn. (CoRL)},
  series = {Proceedings of Machine Learning Research},
  volume = {270},
  pages = {1975--1991},
  year = {2025},
  url = {https://proceedings.mlr.press/v270/zhuang25a.html},
}

@inproceedings{pim,
  author = {Long, Junfeng and Ren, Junli and Shi, Moji and Wang, Zirui and Huang, Tao and Luo, Ping and Pang, Jiangmiao},
  title = {Learning Humanoid Locomotion with Perceptive Internal Model},
  booktitle = {Proc. IEEE Int. Conf. Robot. Autom. (ICRA)},
  pages = {9997--10003},
  year = {2025},
}

@inproceedings{perceptive_humanoid,
  author = {Sun, Wandong and Cao, Baoshi and Chen, Long and Su, Yongbo and Liu, Yang and Xie, Zongwu and Liu, Hong},
  title = {Learning Perceptive Humanoid Locomotion over Challenging Terrain},
  booktitle = {Proc. IEEE/RSJ Int. Conf. Intell. Robots Syst. (IROS)},
  pages = {6571--6578},
  year = {2025},
}

@article{perceptive_wild,
  author = {Miki, Takahiro and Lee, Joonho and Hwangbo, Jemin and Wellhausen, Lorenz and Koltun, Vladlen and Hutter, Marco},
  title = {Learning robust perceptive locomotion for quadrupedal robots in the wild},
  journal = {Sci. Robot.},
  volume = {7},
  number = {62},
  articleno = {eabk2822},
  year = {2022},
  note = {{Art. no.} eabk2822},
}

@inproceedings{extreme_parkour,
  author = {Cheng, Xuxin and Shi, Kexin and Agarwal, Ananye and Pathak, Deepak},
  title = {Extreme Parkour with Legged Robots},
  booktitle = {Proc. IEEE Int. Conf. Robot. Autom. (ICRA)},
  pages = {11443--11450},
  year = {2024},
}

@inproceedings{robot_parkour,
  author = {Zhuang, Ziwen and Fu, Zipeng and Wang, Jianren and Atkeson, Christopher G. and Schwertfeger, S\"{o}ren and Finn, Chelsea and Zhao, Hang},
  title = {Robot Parkour Learning},
  booktitle = {Proc. 7th Conf. Robot Learn. (CoRL)},
  series = {Proceedings of Machine Learning Research},
  volume = {229},
  pages = {73--92},
  year = {2023},
  url = {https://proceedings.mlr.press/v229/zhuang23a.html},
}

@inproceedings{residual_rl,
  author = {Johannink, Tobias and Bahl, Shikhar and Nair, Ashvin and Luo, Jianlan and Kumar, Avinash and Loskyll, Matthias and Ojea, Juan Aparicio and Solowjow, Eugen and Levine, Sergey},
  title = {Residual Reinforcement Learning for Robot Control},
  booktitle = {Proc. IEEE Int. Conf. Robot. Autom. (ICRA)},
  pages = {6023--6029},
  year = {2019},
}

@misc{residual_policy,
  author = {Tom Silver and Kelsey Allen and Josh Tenenbaum and Leslie Kaelbling},
  title = {Residual Policy Learning},
  year = {2018},
  howpublished = {arXiv preprint arXiv:1812.06298},
  eprint = {1812.06298},
  archiveprefix = {arXiv},
  primaryclass = {cs.RO},
  url = {https://arxiv.org/abs/1812.06298},
}

@inproceedings{asymmetric,
  author = {Lerrel Pinto AND Marcin Andrychowicz AND Peter Welinder AND Wojciech Zaremba AND Pieter Abbeel},
  title = {Asymmetric Actor Critic for Image-Based Robot Learning},
  booktitle = {Proc. Robot. Sci. Syst. (RSS)},
  year = {2018},
}

@inproceedings{rma,
  author = {Ashish Kumar AND Zipeng Fu AND Deepak Pathak AND Jitendra Malik},
  title = {{RMA}: Rapid Motor Adaptation for Legged Robots},
  booktitle = {Proc. Robot. Sci. Syst. (RSS)},
  year = {2021},
}

@inproceedings{walk_minutes,
  author = {Rudin, Nikita and Hoeller, David and Reist, Philipp and Hutter, Marco},
  title = {Learning to Walk in Minutes Using Massively Parallel Deep Reinforcement Learning},
  booktitle = {Proc. 5th Conf. Robot Learn. (CoRL)},
  series = {Proceedings of Machine Learning Research},
  volume = {164},
  pages = {91--100},
  year = {2022},
  url = {https://proceedings.mlr.press/v164/rudin22a.html},
}

@misc{humanoid_gym,
  author = {Xinyang Gu and Yen-Jen Wang and Jianyu Chen},
  title = {{Humanoid-Gym}: Reinforcement Learning for Humanoid Robot with Zero-Shot {Sim2Real} Transfer},
  year = {2024},
  howpublished = {arXiv preprint arXiv:2404.05695},
  eprint = {2404.05695},
  archiveprefix = {arXiv},
  primaryclass = {cs.RO},
  url = {https://arxiv.org/abs/2404.05695},
}

@inproceedings{videomimic,
  author = {Allshire, Arthur and Choi, Hongsuk and Zhang, Junyi and McAllister, David and Zhang, Anthony and Kim, Chung Min and Darrell, Trevor and Abbeel, Pieter and Malik, Jitendra and Kanazawa, Angjoo},
  title = {Visual Imitation Enables Contextual Humanoid Control},
  booktitle = {Proc. 9th Conf. Robot Learn. (CoRL)},
  series = {Proceedings of Machine Learning Research},
  volume = {305},
  pages = {794--815},
  year = {2025},
  url = {https://proceedings.mlr.press/v305/allshire25a.html},
}

@misc{isaaclab,
  author = {Mayank Mittal and Pascal Roth and James Tigue and Antoine Richard and Octi Zhang and Peter Du and Antonio Serrano-Muñoz and Xinjie Yao and René Zurbrügg and Nikita Rudin and Lukasz Wawrzyniak and Milad Rakhsha and Alain Denzler and Eric Heiden and Ales Borovicka and Ossama Ahmed and Iretiayo Akinola and Abrar Anwar and Mark T. Carlson and Ji Yuan Feng and Animesh Garg and Renato Gasoto and Lionel Gulich and Yijie Guo and M. Gussert and Alex Hansen and Mihir Kulkarni and Chenran Li and Wei Liu and Viktor Makoviychuk and Grzegorz Malczyk and Hammad Mazhar and Masoud Moghani and Adithyavairavan Murali and Michael Noseworthy and Alexander Poddubny and Nathan Ratliff and Welf Rehberg and Clemens Schwarke and Ritvik Singh and James Latham Smith and Bingjie Tang and Ruchik Thaker and Matthew Trepte and Karl Van Wyk and Fangzhou Yu and Alex Millane and Vikram Ramasamy and Remo Steiner and Sangeeta Subramanian and Clemens Volk and CY Chen and Neel Jawale and Ashwin Varghese Kuruttukulam and Michael A. Lin and Ajay Mandlekar and Karsten Patzwaldt and John Welsh and Huihua Zhao and Fatima Anes and Jean-Francois Lafleche and Nicolas Moënne-Loccoz and Soowan Park and Rob Stepinski and Dirk Van Gelder and Chris Amevor and Jan Carius and Jumyung Chang and Anka He Chen and Pablo de Heras Ciechomski and Gilles Daviet and Mohammad Mohajerani and Julia von Muralt and Viktor Reutskyy and Michael Sauter and Simon Schirm and Eric L. Shi and Pierre Terdiman and Kenny Vilella and Tobias Widmer and Gordon Yeoman and Tiffany Chen and Sergey Grizan and Cathy Li and Lotus Li and Connor Smith and Rafael Wiltz and Kostas Alexis and Yan Chang and David Chu and Linxi "Jim" Fan and Farbod Farshidian and Ankur Handa and Spencer Huang and Marco Hutter and Yashraj Narang and Soha Pouya and Shiwei Sheng and Yuke Zhu and Miles Macklin and Adam Moravanszky and Philipp Reist and Yunrong Guo and David Hoeller and Gavriel State},
  title = {{Isaac Lab}: A {GPU}-Accelerated Simulation Framework for Multi-Modal Robot Learning},
  year = {2025},
  howpublished = {arXiv preprint arXiv:2511.04831},
  url = {https://arxiv.org/abs/2511.04831},
}

@inproceedings{attention,
  author = {Vaswani, Ashish and Shazeer, Noam and Parmar, Niki and Uszkoreit, Jakob and Jones, Llion and Gomez, Aidan N and Kaiser, \L ukasz and Polosukhin, Illia},
  title = {Attention is All you Need},
  booktitle = {Adv. Neural Inf. Process. Syst.},
  volume = {30},
  year = {2017},
  url = {https://proceedings.neurips.cc/paper_files/paper/2017/file/3f5ee243547dee91fbd053c1c4a845aa-Paper.pdf},
}

@misc{pgmt,
  author = {Li, Hongyi and Li, Peizhuo and Tao, Yucheng and Wang, Ze and Xu, Fangzhou and Chen, Jinyi and Yuan, Yanyan and Jia, Dapeng and Jin, Yongbin and Fan, Mingfeng and Sartoretti, Guillaume and Wang, Hongtao},
  title = {{PGMT}: Perceptive General Motion Tracking for Humanoid Robots},
  year = {2026},
  howpublished = {arXiv preprint arXiv:2609.08511},
  eprint = {2609.08511},
  archiveprefix = {arXiv},
  primaryclass = {cs.RO},
  url = {https://arxiv.org/abs/2609.08511},
}

@inproceedings{perceptive_bfm,
  author = {Zifan Wang and Yizhao Li and Teli Ma and Qiang Zhang and Yudong Fan and Hao Xu and Shuo Yang and Junwei Liang},
  title = {Perceptive Behavior Foundation Model: Adapting Human Motion Priors to Robot-Centric Terrain},
  booktitle = {Proc. 10th Conf. Robot Learn. (CoRL)},
  year = {2026},
  url = {https://arxiv.org/abs/2606.08059},
}

@inproceedings{pmlr-v15-ross11a,
  author = {Ross, Stephane and Gordon, Geoffrey and Bagnell, Drew},
  title = {A Reduction of Imitation Learning and Structured Prediction to No-Regret Online Learning},
  booktitle = {Proc. 14th Int. Conf. Artif. Intell. Stat. (AISTATS)},
  series = {Proceedings of Machine Learning Research},
  volume = {15},
  pages = {627--635},
  year = {2011},
  url = {https://proceedings.mlr.press/v15/ross11a.html},
}

@inproceedings{AMASS:ICCV:2019,
  author = {Mahmood, Naureen and Ghorbani, Nima and Troje, Nikolaus F. and Pons-Moll, Gerard and Black, Michael J.},
  title = {{AMASS}: Archive of Motion Capture as Surface Shapes},
  booktitle = {Proc. IEEE/CVF Int. Conf. Comput. Vis. (ICCV)},
  pages = {5441--5450},
  year = {2019},
}

@misc{bones_seed,
  author = {{Bones Studio}},
  title = {{BONES-SEED}: Skeletal Everyday Embodiment Dataset},
  howpublished = {Hugging Face Dataset},
  url = {https://huggingface.co/datasets/bones-studio/seed},
  note = {Accessed: 2026},
}

@misc{motiondecode,
  author = {{ChingMu}},
  title = {{MotionDecode}: {ChingMu} 1000-Hour Embodied Motion Dataset},
  year = {2026},
  howpublished = {Hugging Face Dataset},
  url = {https://huggingface.co/datasets/CMRobot/MotionDecode},
  note = {Accessed: 2026},
}

@IEEEtranBSTCTL{bibcontrol,
  CTLuse_forced_etal = {yes},
  CTLmax_names_forced_etal = {6},
  CTLnames_show_etal = {1},
  CTLdash_repeated_names = {no}
}
\clearpage
\raggedcolsend
\flushbottom
\appendices
% Appendix starts on a fresh page after all references. A full-width block keeps
% the two configuration tables and the reward table together, without floats.
\onecolumn
\newsavebox{\compactappendixbox}
\sbox{\compactappendixbox}{%
\begin{minipage}{\textwidth}
\section{Implementation Details}
\label{sec:architecture}
\noindent\small
Convolutional neural networks (CNNs) use stride-2 $3\times3$ kernels and $4\times4$ pooling; attention has four heads.
MLP denotes a multilayer perceptron. Both stages use Adam with 24-step rollouts and gradient clipping at 1.
The teacher uses 12 curriculum rows, with $\ell_t=\operatorname{clip}(k_t/11,0,1)$ for raw row index $k_t$. Rewards use $r=0.02\sum_jw_jr_j$, with FI/UW regularization applied separately.
\par\vspace{5pt}
\expandafter\def\csname @captype\endcsname{table}
\noindent\begin{minipage}[t]{0.485\textwidth}
\vspace{0pt}
% Nonfloating fragment used on compact appendix page.
\centering
\caption{Residual-policy architecture.}
\label{tab:actor_architecture}
\fontsize{8.5}{9.5}\selectfont
\setlength{\tabcolsep}{3pt}
\begin{tabular*}{\linewidth}{@{\extracolsep{\fill}}ll@{}}
\toprule
Component & Configuration \\
\midrule
\multicolumn{2}{@{}l}{\textit{Common layer widths}} \\
Motion / action projection & $64\to256$ / $29\to256$ \\
Proprioceptive MLP & $930\to512\to256$ \\
Residual actor hidden widths & $2048,1024,1024$ \\
\midrule
\multicolumn{2}{@{}l}{\textit{Privileged teacher (ours)}} \\
Teacher height CNN & $1\to32\to64$ \\
Foot-query MLP & $50\to128\to64$ \\
Teacher attention / context & $64$ / $64\to256$ \\
Fusion gate & $128\to64\to2$ \\
Critic hidden widths & $256,256,256$ \\
\midrule
\multicolumn{2}{@{}l}{\textit{Depth student (ours)}} \\
Student depth CNN & $10\to32\to64$ \\
Student token projection & $64\to256$ \\
\bottomrule
\end{tabular*}

\end{minipage}\hfill
\begin{minipage}[t]{0.485\textwidth}
\vspace{0pt}
% Nonfloating fragment used on compact appendix page.
\centering
\caption{PPO and DAgger settings.}
\label{tab:training}
\fontsize{8.5}{9.5}\selectfont
\renewcommand{\arraystretch}{1.04}
\setlength{\tabcolsep}{4pt}
\begin{tabular*}{\linewidth}{@{\extracolsep{\fill}}lr@{}}
\toprule
Setting & Value \\
\midrule
\multicolumn{2}{@{}l}{\textit{Privileged teacher (ours): PPO}} \\
Parallel environments & 16,384 \\
Epochs / minibatches & 5 / 4 \\
Actor learning rate (initial) & $10^{-4}$ \\
Adaptive LR range & $[10^{-5},2\!\times\!10^{-4}]$ \\
Target KL / critic learning rate & $0.01$ / $10^{-3}$ \\
Discount / GAE $\lambda$ & 0.99 / 0.95 \\
PPO clip / entropy coefficient & 0.2 / 0.0015 \\
\midrule
\multicolumn{2}{@{}l}{\textit{Depth student (ours): DAgger}} \\
Parallel environments & 8,192 \\
Epochs / minibatches & 1 / 4 \\
Learning rate & $5\!\times\!10^{-4}$ \\
Huber threshold / token coefficient & 1 / 0.5 \\
\bottomrule
\end{tabular*}

\end{minipage}
\par\vspace{7pt}
% Generated by scripts/prepare_reward_table.py.
% Full-width nonfloating fragment for compact appendix.
\centering
\caption{Teacher rewards and signed weights. $\mathcal G(e;\sigma)=\exp[-(e/\sigma)^2]$.}
\label{tab:teacher_rewards}
\fontsize{8.5}{9.5}\selectfont
\renewcommand{\arraystretch}{1.0}
\setlength{\tabcolsep}{4pt}
\begin{tabular*}{\linewidth}{@{\extracolsep{\fill}}p{0.25\linewidth}p{0.61\linewidth}r@{}}
\toprule
Term & Raw reward / penalty & Weight \\
\midrule
\multicolumn{3}{@{}l}{\textit{Motion tracking ($r_{\mathrm{track}}$)}} \\
Root XY position & $\mathcal G(e_{xy};0.3)$ & $0.5$ \\[0.2pt]
Root height above feet & $\mathcal G(e_z;0.5)$ & $0.5$ \\[0.2pt]
Root planar velocity & $\mathcal G(e_{v,xy};1.0)$ & $1$ \\[0.2pt]
Root orientation & $\mathcal G(e_R;0.4)$ & $0.5$ \\[0.2pt]
Upper-body position & $\mathcal G(e_{\mathcal U}^{p};0.3)$ & $1$ \\[0.2pt]
Upper-body orientation & $\mathcal G(e_{\mathcal U}^{R};0.4)$ & $1$ \\[0.2pt]
Upper-body linear velocity & $\mathcal G(e_{\mathcal U}^{v};1.0)$ & $0.5$ \\[0.2pt]
Upper-body angular velocity & $\mathcal G(e_{\mathcal U}^{\omega};3.14)$ & $0.5$ \\[0.2pt]
Lower-body position & $\mathcal G([e_{\mathcal L}^p-\delta_p(\ell_t)]_+;0.45)$ & $0.5$ \\[0.2pt]
Lower-body orientation & $\mathcal G([e_{\mathcal L}^R-\delta_R(\ell_t)]_+;0.5)$ & $1$ \\[0.2pt]
Lower-body linear velocity & $\mathcal G(e_{\mathcal L}^{v};1.5)$ & $0.5$ \\[0.2pt]
Lower-body angular velocity & $\mathcal G(e_{\mathcal L}^{\omega};3.925)$ & $0.5$ \\[0.2pt]
\midrule
\multicolumn{3}{@{}l}{\textit{Contact and terrain traversal ($r_{\mathrm{adapt}}$)}} \\
Reference-stance support & $\frac{\sum_i\kappa_{i,t}\rho_{i,t}\alpha_{i,t}\eta_{i,t}}{\sum_i\kappa_{i,t}\rho_{i,t}}$ & $2$ \\[0.2pt]
Swept clearance & $\tfrac12\sum_i b_{i,t}^{\mathrm{sw}}\mathcal G([d_{\mathrm{clr}}-c_{i,t}]_+;\sigma_{\mathrm{clr}})$ & $2$ \\[0.2pt]
Out-of-window contact & $\tfrac12\sum_i\kappa_{i,t}\alpha_{i,t}(1-\rho_{i,t})$ & $-3$ \\[0.2pt]
Contact duration shortfall & $\tfrac12\sum_i\kappa_{i,t}b_{i,t}^{\mathrm{end}}[0.7-\beta_{i,t}]_+$ & $-6$ \\[0.2pt]
Repeated touchdown & $\tfrac12\sum_i\kappa_{i,t}b_{i,t}^{\mathrm{on}}\mathbf1[N_{i,t}^{-}\geq1]$ & $-12$ \\[0.2pt]
Unsupported touchdown & $\sum_i b_{i,t}^{\mathrm{td}}\mathbf1[T_{i,t}^{\mathrm{air}}\geq0.08\ \land\ \eta_{i,t}^{\mathrm{td}}<0.9]$ & $-5$ \\[0.2pt]
Stance slip & $\min(1,\tfrac12\sum_i\chi_{i,t}\|v_{i,xy}\|^2/0.25^2)$ & $-2$ \\[0.2pt]
Stumble & $\sum_i\mathbf1[\|F_{i,xy}\|>3|F_{i,z}|\ \land\ \|F_{i,xy}\|>1\,\mathrm N]$ & $-15$ \\[0.2pt]
Non-foot contact & $\sum_{b\in\mathcal B_{\mathrm{nf}}}\mathbf1[F_{b,t}^{\max}>20\,\mathrm N]$ & $-3$ \\[0.2pt]
Reference-stance indicator & $\mathbf1[\max_i\rho_{i,t}>0.5]$ & $10^{-6}$ \\[0.2pt]
\midrule
\multicolumn{3}{@{}l}{\textit{Motion regularization and safety ($r_{\mathrm{reg}}$)}} \\
Action rate & $\|a_t-a_{t-1}\|_2^2$ & $-0.05$ \\[0.2pt]
Residual rate at low speed & $\mathbf1[\|v_{xy}^{\mathrm{ref}}\|<0.1\,\mathrm{m/s}]\|\Delta a_t-\Delta a_{t-1}\|_2^2$ & $-3$ \\[0.2pt]
Joint-position limits & $\sum_j([q_j-q_j^{\max}]_+ +[q_j^{\min}-q_j]_+)$ & $-10$ \\[0.2pt]
Wrist/head angular velocity & $\langle[\|\omega_b\|-1.5]_+^2\rangle_{b\in\mathcal W}$ & $-0.005$ \\[0.2pt]
Ankle acceleration & $(1-0.75\ell_t)\sum_{j\in\mathcal A}\ddot q_j^2$ & $-2.5\!\times\!10^{-6}$ \\[0.2pt]
Hard termination & $\mathbf1[\text{hard termination}]$ & $-30$ \\[0.2pt]
\bottomrule
\end{tabular*}
\vspace{3pt}
\begin{minipage}{\linewidth}\footnotesize
$e$: tracking error (SI units); $\mathcal U/\mathcal L$: upper/lower bodies; $\delta_p(\ell_t)=0.02+0.08\ell_t$\,m, $\delta_R(\ell_t)=0.10+0.20\ell_t$\,rad; $b^{\mathrm{end/on/td}}$: interval-end/onset/touchdown indicators; $\beta$: contact coverage; $N^{-}$: prior onsets; $\chi$: actual stance; $T^{\mathrm{air}}$: air time; $\eta^{\mathrm{td}}$: sole support; $F^{\max}$: peak contact force; $\mathcal B_{\mathrm{nf}},\mathcal A,\mathcal W$: non-foot bodies, ankle joints, wrists/head; $q^{\min/\max}$: soft joint limits; $\langle\cdot\rangle$: arithmetic mean. Support is zero if its denominator is zero.
\end{minipage}

\end{minipage}%
}
\typeout{COMPACT APPENDIX HEIGHT: \the\dimexpr\ht\compactappendixbox+\dp\compactappendixbox\relax; TEXT HEIGHT: \the\textheight}
\noindent\usebox{\compactappendixbox}

\end{document}